\documentclass{bmvc2k}

\usepackage[utf8]{inputenc} 
\usepackage[T1]{fontenc}    
\usepackage{hyperref}       
\usepackage{url}            
\usepackage{booktabs}       
\usepackage{amsfonts}       
\usepackage{nicefrac}       
\usepackage{microtype}      
\usepackage{graphicx}
\usepackage{epstopdf}
\usepackage{subfigure}
\usepackage{enumitem}

\title{A Spatial and Temporal Features Mixture Model with Body Parts for Video-based Person Re-Identification}

\addauthor{Jie Liu}{}{1}
\addauthor{Cheng Sun}{}{1}
\addauthor{Xiang Xu}{}{2}
\addauthor{Baomin Xu}{bmxu@bjtu.edu.cn}{1}
\addauthor{Shuangyuan Yu}{}{1}

\addinstitution{
 School of Computer and Information Technology\\
 Beijing Jiaotong University\\
 Beijing 100044, P. R. China
}
\addinstitution{
 Department of Electrical and Computer Engineering\\
 Carnegie Mellon University\\
 PA 15213, USA
}

\runninghead{Student, Prof, Collaborator}{BMVC Author Guidelines}

\def\etal{\emph{et al}\bmvaOneDot}

\begin{document}

\maketitle

\begin{abstract}
  The video-based person re-identification is to recognize a person under different cameras, which is a crucial task applied in visual surveillance system. Most previous methods mainly focused on the feature of full body in the frame. In this paper we propose a novel Spatial and Temporal Features Mixture Model (STFMM) based on convolutional neural network (CNN) and recurrent neural network (RNN), in which the human body is split into $N$ parts in horizontal direction so that we can obtain more specific features. The proposed method skillfully integrates features of each part to achieve more expressive representation of each person. We first split the video sequence into $N$ part sequences which include the information of head, waist, legs and so on. Then the features are extracted by STFMM whose $2N$ inputs are obtained from the developed Siamese network, and these features are combined into a discriminative representation for one person. Experiments are conducted on the iLIDS-VID and PRID-2011 datasets. The results demonstrate that our approach outperforms existing methods for video-based person re-identification. It achieves a rank-1 CMC accuracy of 74\% on the iLIDS-VID dataset, exceeding the the most recently developed method ASTPN by 12\%. For the cross-data testing, our method achieves a rank-1 CMC accuracy of 48\% exceeding the ASTPN method by 18\%, which shows that our model has significant stability.
\end{abstract}

\section{Introduction}
\label{sec:intro}
Person re-identification task aims to track a person appeared in non-overlapping cameras at distinct times \cite{bt1}. It has drawn much attention due to the huge demand as its applications in security and surveillance domain. In real-world system, it is still a challenging problem because different images may contain variations of occlusions, complicated backgrounds, illuminations and view points.

Methods of person re-identification in single still images have been widely investigated and have actively promoted the development of this topic. These methods mainly consist of two aspects: feature learning \cite{bt7,bt15,bt16,bt17,bt18,bt22,bt34} and metric learning \cite{bt5,bt6,bt19,bt20,bt21,bt23,bt24,bt25,bt26}. Feature learning mainly focuses on extracting discriminative features and building an invariant representation for each person, such as pose and clothes color information. On the other hand, metric learning aims to minimize the variance of same person and maximize that of the different ones. Compared with single image, video sequence are closer to the real scenario which can be captured by a surveillance camera. Obviously, video sequence inherently carries plentiful information. Person's motion, such as gait, can be extracted from the temporal series which help build unique features of one person. However, a lot of redundant information is also contained in the video, such as multifarious background and occlusion, which makes the feature extracting more difficult.

\begin{figure}
  \centering
  \includegraphics[width=0.8\textwidth]{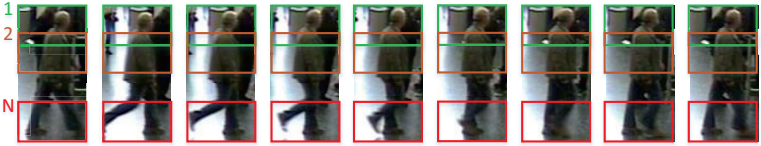}\\
  \caption{Each sequence is split into $N$ continuous part sequences, including head part (green box), shoulder part (brown box), $\ldots$, and shank part (red box). }
  \label{figure1}
\end{figure}

Recently, more and more researchers are interested in video-based person re-identification \cite{bt3,bt8,bt9,bt10,bt11,bt12,bt13}. As deep neural network (DNN) moves ahead in generic object recognition, DNN-based schemes have also been successfully applied in this task. Given a video sequence, convolutional neural network (CNN) works as a feature extractor for each frame, while recurrent neural network (RNN) exploits the temporal information from the video sequence. Siamese network \cite{bt14} is a popular deep neural network architecture in person recognition tasks, with which a lot of methods based on CNN-RNN-scheme have achieved reasonable successes \cite{bt2,bt8,bt12}. Since the mode of CNN-RNN-scheme based on Siamese is so popular, we adopt it for video-based person re-identification.

However, most previous methods based on CNN and RNN simply extract full body features and create sequence-level representation of each person, in which the features of the body parts are not considered specially. In this paper, we propose a novel Spatial and Temporal Features Mixture Model (STFMM) for extracting the body parts information, as shown in Fig. \ref{figure1}. Given a pair of video sequences, we first split each frame into $N$ parts, and the adjacent parts are overlapped with several pixels. Next we utilize STFMM to process part sequences concurrently and calculate the similarity of two corresponding features of each part. Finally, we propose an algorithm to mix the $N$ part features to generate the discriminative representation of each person. The extensive experiments are carried out on two datasets, iLIDS-VID and PRID-2011. The experimental results demonstrate that our approach outperforms existing methods for video-based person re-identification. It achieves 74\% for rank-1 matching rate on the iLIDS-VID dataset, exceeding the the most recently developed method ASTPN by 12\%. For the cross-data testing, our method achieves 48\% exceeding the ASTPN method by 18\%, which shows that our model has significant stability.

We summarize the contributions of this work in three folds as follows,

(1) In order to improve the accuracy of re-identification, we propose a novel method of Spatial and Temporal Features Mixture Model (STFMM) by making full use of the information of each human part. Using the STFMM, we can extract the features of each part sequence respectively, and combine these features into the discriminative representation of one person.

(2) To satisfy the input of our network, we develop the Siamese network architecture in which the number of input are adjusted from $2$ to $2N$.

(3) By combining the STFMM and developed Siamese network, our method achieves better results and has more significant stability than the state-of-the-art methods.
\section{Related Work}
\label{relatedwork}

Traditional methods for person re-identification mainly consist of two aspects: feature learning and metric learning. Many feature learning methods have been proposed for person re-identification in single still images or video sequences. Matsukawa \etal \cite{bt15} presented a descriptor on a hierarchical distribution of pixel features and used Gaussian distribution to describe a local image region. Liao \etal \cite{bt34} utilized the horizontal occurrence of local features and maximized the occurrence for stable representation, called Local Maximal Occurrence (LOMO). Wang \etal \cite{bt10} proposed a Discriminative Video Ranking model (DVR) to select the most discriminative video fragments, from which more reliable space-time features can be extracted. The method of Bag-of-Words (BoW) \cite{bt28} aimed to learn a mapping function that converted frame-wise features to a global vector. The metric learning methods have also been widely invested and made some positive achievements, such as Relaxed Pairwise Learning (RPL) \cite{bt23}, Large Margin Nearest-Neighbour (LMNN) \cite{bt24}, Relevance Component Analysis (RCA) \cite{bt25}, Locally Adaptive Decision Function (LADF) \cite{bt26}, and RankSVM \cite{bt7}.

Deep neural network (DNN) has achieved significant successes in computer vision, and the DNN-based methods have been studied and applied in person re-identification task \cite{bt3,bt4,bt8,bt10,bt12,bt13,bt27,bt28}. The DNN-based models are trained with a pair of inputs for learning a direct mapping from image or video sequence to feature space. Mclaughlin \etal \cite{bt8} combined CNN, RNN and Siamese network together, which is the first time applying DNN to the video-based re-identification. Attention mechanism \cite{bt29,bt30} has gained huge achievement in deep learning. Xu \etal \cite{bt12} proposed a joint Spatial and Temporal Attention Pooling Network (ASTPN) to extract sequence-level features by selecting informative frames and notable regions of each frame. Zhou \etal \cite{bt13} used Temporal Attention Model (TAM) to measure the importance of each frame in video sequence and applied Spatial Recurrent Model (SRM) to explore contextual information.

However, most of CNN-RNN-based methods mainly pay attention to extract the feature of full body for creating sequence-level representation. In order to utilize the information of body parts, in this paper, we propose a novel Spatial and Temporal Features Mixture Model (STFMM) for learning the significant representation consisting of the features of part sequences.
\begin{figure}
  \centering
  \includegraphics[width=0.8\textwidth]{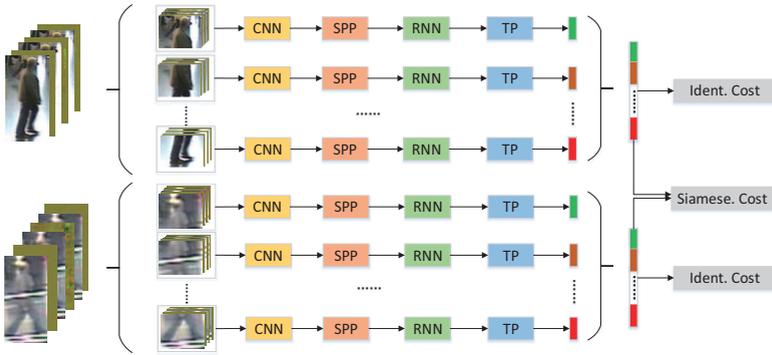}\\
  \caption{The architecture of STFMM.}
  \label{figure2}
\end{figure}
\section{Method}

The architecture we proposed is shown in Fig. \ref{figure2}. In this architecture each video sequence is first split into $N$ part sequences as the input of STFMM. Convolutional neural network (CNN) works as a feature extractor, after which we employ the spatial pyramid pooling (SPP) \cite{bt31} to generate the image-level features. Then these feature sequences are fed into recurrent neural network (RNN), which is a powerful model to deal with the temporal sequences. After RNN, we adopt the method of Temporal Pooling (TP) \cite{bt8} to average the spatial features over time-steps. Finally, the part-level features are combined by an algorithm for mixing features to form the sequence-level representation.

In order to satisfy the input of our network STFMM, we develop the Siamese network architecture \cite{bt14}. Different from the original Siamese network, the number of input is adjusted from $2$ to $2N$. 
In the following subsections we will explain about each components of STFMM in detail.

\subsection{Input}
\label{netinput}

The input data of the network consists of optical flow and color information. Optical flow consists of the horizontal and vertical channels, which is the pattern of apparent motion of image objects between two consecutive frames. Color information has three color channels that encode the information of person's appearance. Many researchers \cite{bt8,bt12,bt13} show that using both optical flow and color channels as the input of the network can improve the accuracy of person re-identification. In our method, we split each frame of video sequence into $N$ parts, in which each part frame consists of five channels, two being for optical flow and three being for color information.

Given a video sequence $\mathcal{S}=\{s^1,s^2,\ldots,s^T\}$, where $T$ is the sequence length and $s^t$ represents the frame of time $t$, $1\leq t\leq T$. Assuming that the parts number is $N$, and the number of overlapping pixels is $p$, the height of each part can be calculated as $h=\big\lfloor\frac{1}{N}(H+(N-1)p)\big\rfloor$,
where $H$ is the height of original frame. Let $\mathcal{S}_n=\{s_n^1,s_n^2,\ldots,s_n^T\}$ be the $n$-th part sequence, $1\leq n \leq N$, the input of our network is defined as $\mathcal{V}=\{\mathcal{S}_1,\mathcal{S}_2,\ldots,\mathcal{S}_N\}$.

\subsection{Convolutional Layer}

In our architecture, CNN consists of three convolutional layers $l_1, l_2$ and $l_3$, in which $l_1$ and $l_2$ consist of convolution, non-linear activation-function ($\tanh$) and max-pooling. We define $C_{l_i}(s_n^t)$ as the operation function of $i$-th convolutional layers for each part sequence, then $C_{l_1}(s_n^t)=Maxpool(\tanh(Conv(s_n^t)))$ and $C_{l_2}(O_1)=Maxpool(\tanh(Conv(O_1)))$,
where $s_n^t$ is the frame of $n$-th part sequence at time $t$, and $O_1=C_{l_1}(s_n^t)$ is the output of $l_1$ layer. For the last convolutional layer $l_3$, similar to \cite{bt12}, we employ the spatial pyramid pooling (SPP) to replace the max-pooling,
\begin{equation}
  C_{l_3}(O_2)=Spp(\tanh(Conv(O_2))),
  \label{thirdconv}
\end{equation}
where $Spp$ represents the spatial pyramid pooling operation and $O_2=C_{l_2}(O_1)$ is the output of $l_2$ layer. Now we consider the operation in Eq. \ref{thirdconv}. Let ${C_{l_3}}'$ be the output of $\tanh(Conv(O_2))$, 
then ${C_{l_3}}'=\{{C_{l_3,1}}',{C_{l_3,2}}',\ldots,{C_{l_3,N}}'\}$ and ${C_{l_3,n}}'=\{{c_{l_3,n}^1}',{c_{l_3,n}^2}',\ldots,{c_{l_3,n}^T}'\}$,
where ${c_{l_3,n}^t}'\in\mathbb{R}^{c\times w\times h}$ is the output of $\tanh(Conv(O_2))$ for the $n$-th part sequence at time $t$, $c$ is the output channels, $w$ and $h$ are the width and height of each feature map respectively.


Now we discuss the SPP in our architecture. Let $e$ be the number of spatial bins, and the size of spatial bins $B_j=\{(b_j,b_j)|1\leq j\leq e\}$. In our model,we use SPP to generate a fixed-length representation with multi-level spatial bins $8\times8$, $4\times4$, $2\times2$ and $1\times1$. 
We define $\mathcal{SPP}$ as the comprehensive image-level features extracted by the spatial pooling layer, then $\mathcal{SPP}=\{\mathcal{SPP}_1,\mathcal{SPP}_2,\ldots,\mathcal{SPP}_N\}$ and $\mathcal{SPP}_n=\{spp_n^1,spp_n^2,\ldots,spp_n^T\}$,
where $spp_n^t\in \mathbb{R}^{q}$ $(q=\sum_{j=1}^eb_jb_j)$ is the output of $Spp({c_{l_3,n}^t}')$ for the $n$-th part sequence at time $t$. We use window size $win_j=(\lceil\frac{w}{b_j}\rceil,\lceil\frac{h}{b_j}\rceil)$ and pooling stride $str_j=(\lfloor\frac{w}{b_j}\rfloor,\lfloor\frac{h}{b_j}\rfloor)$ to perform the pyramid pooling. Each $spp_n^t$ can be calculate as follows:
\begin{equation}
  spp_n^t=bin_{n,1}^t\oplus bin_{n,2}^t\oplus\ldots\oplus bin_{n,e}^t,
\end{equation}
\begin{equation}
  bin_{n,j}^t=F(M({c_{l_3,n}^t}';win_j,str_j)),
\end{equation}
where the function $M$ performs max pooling on ${c_{l_3,n}^t}'$ with the sizes of $win_j$ and $str_j$, the function $F$ turns the output of max pooling into a one-dimensional vectors, and the operator $\oplus$ joints these vectors together. After the spatial pyramid pooling layer, the part features $\mathcal{SPP}$ will be passed on to the recurrent network, see the next section in detail.

\subsection{Recurrent Layer}
Recurrent neural network (RNN) is able to extract feature of temporal sequence, which help produce an output based on both current input and previous information at each time-step. We employ RNN to handle the part features $\mathcal{SPP}$, aiming to capture the temporal information of each sequence. The recurrent layer can be formulized as follows:
\begin{equation}
  o_n^t=Uspp_n^t+Wr_n^{t-1}, \ \ \ \
\end{equation}
\begin{equation}
  r_n^t=\tanh(o_n^t),
\end{equation}
where $U$ and $W$ are the parameters of RNN unit, $o_n^t$ represents a linear combination of the current input $spp_n^t$ and $r_n^{t-1}$ which is the previous information of the RNN's state at time $t-1$.

The final output of RNN is greatly affected by the later time-steps. However, the notable frames could appear in any place of the sequence. To solve these problems, similar to \cite{bt8}, we employ the temporal pooling (TP) to capture the long-term information, and adopt mean-pooling over the temporal dimension to extract a single feature vector of each part. The TP layer can be formulized as,
\begin{equation}
  tp_n=\frac{1}{T}\sum_{t=1}^{T}o_n^t.
\end{equation}
After the TP layer, we get the part-level features of one video $\mathcal{T}=\{tp_1,tp_2,\ldots,tp_N\}$. In order to comprehensively extract the representation of video sequence from these part-level features, we propose an algorithm to mix features in $\mathcal{T}$ as follows:
\begin{equation}
  v_p=tp_1^p\oplus tp_2^p\oplus\ldots\oplus tp_N^p, \ \ \ \
  v_g=tp_1^g\oplus tp_2^g\oplus\ldots\oplus tp_N^g,
\end{equation}
where the operator $\oplus$ joints the part-level features together, $tp_n^p$ and $tp_n^g$ represent the $n$-th part features of the probe and gallery video sequences respectively, $v_p$ and $v_g$ represent the final representation of the probe and gallery video sequences.

\subsection{Loss Function}
\label{lossfunction}

Similar to \cite{bt12}, we train STFMM with three loss functions, including two person's identification losses and Siamese loss. Given the probe and the gallery video sequences, we utilize STFMM to capture the discriminative representations $v_p$ and $v_g$. $I(v_p)$ and $I(v_g)$ are defined as the person's identification losses, which employ softmax regression and the standard cross-entropy loss. The Siamese loss of two sequences $E(v_p,v_g)$ is defined as follows:
\begin{equation} 
  E(v_p,v_g)=
  \left\{
  \begin{array}{ll}
  \sum_{n=1}^{N}\|tp_{n}^{p}-tp_{n}^{g}\|^{2}, & p=g,\\
  \max\{0,m-\sum_{n=1}^{N}\|tp_{n}^{p}-tp_{n}^{g}\|^{2}\}, & p\neq g,
  \end{array}
  \right.
  \label{marginequation}
\end{equation}
where $\|\cdot\|^2$ is the Euclidean distance of two vectors, $m$ is a margin to separate features of different people, and its value will be discussed in section \ref{npartmarginoverlapping}. We define the overall training objective as $L(v_p,v_g)$, which combines the person's identification losses and Siamese loss, 
\begin{equation}
  L(v_p,v_g)=I(v_p)+I(v_g)+E(v_p,v_g).
\end{equation}

\section{Experiments}

\subsection{Data Preparation and Experiment Settings}
\begin{figure}
  \centering
  \includegraphics[width=0.6\textwidth]{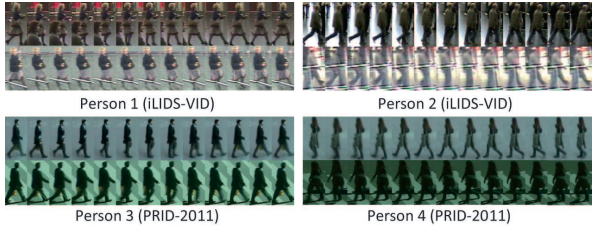}\\
  \caption{The samples of iLIDS-VID (first row) and PRID-2011 (second row).}
  \label{figure4}
\end{figure}

The iLIDS-VID dataset \cite{bt10} contains a total of 300 pairs of video sequences, which are captured by two non-overlapping cameras at an airport arrival hall under CCTV networks. Each person is represented by two video sequences with the lengths ranging from 23 to 192 frames. The PRID-2011 dataset \cite{bt32} consists of 749 persons in which each person is captured by two non-overlapping cameras, and the lengths of sequences range from 5 to 675 frames. We only use the first 200 persons in PRID-2011 dataset who appear in both cameras. We notice that the video sequences in iLIDS-VID are more challenging than that in PRID-2011, for example we select 4 video sequences randomly as shown in Fig. \ref{figure4}.

To train and test the proposed network, we evenly split each dataset into two subsets randomly, in which one was for training and the other was for testing. We repeated experiments 10 times with different train/test splits, and calculated the average of the results to get stable results. Since the deep neural network required a large amount of data during the process of training, we did data augmentation by random mirroring and cropping on the part sequences to increase the diversities of data.

Similar to \cite{bt8, bt12}, we randomly chose sub-sequence of $k=16$ consecutive frames from probe and gallery datasets for the training at each epoch. A full epoch consisted of all positive pairs and same amount of negative pairs. In each epoch, the positive pairs and negative pairs were used alternately. The positive pairs consisted of the two sub-sequences of the same person A from camera 1 and camera 2, and the negative pairs consisted of two sub-sequences of person A and B captured from two cameras respectively. During the testing process, we treated the video sequences of the camera 1 as the probe sets and the camera 2 as the gallery sets, where the data augmentation operation was also applied to all image sequences.

Before being passed on to the network, video sequences were converted to YUV color space and each color channel was normalized to have zero mean and unit variance. We used the Lucas-Kanade method \cite{bt33} to calculate the horizontal and vertical optical flow channels. Then optical flow channels were normalized to the range from -1 to 1. Both optical flow and color information were used as input data, which consisted of five channels in training and testing processes.

We trained the network by using the stochastic gradient descent with batch size of one and learning rate of $1e-3$. Our model was trained for 700 epochs by using a GPU of Nvidia GTX-1080.
\begin{figure}
  \centering
  \subfigure[]{
    \begin{minipage}{0.48\textwidth}
    \centering
    \includegraphics[width=0.9\textwidth, height=0.7\textwidth]{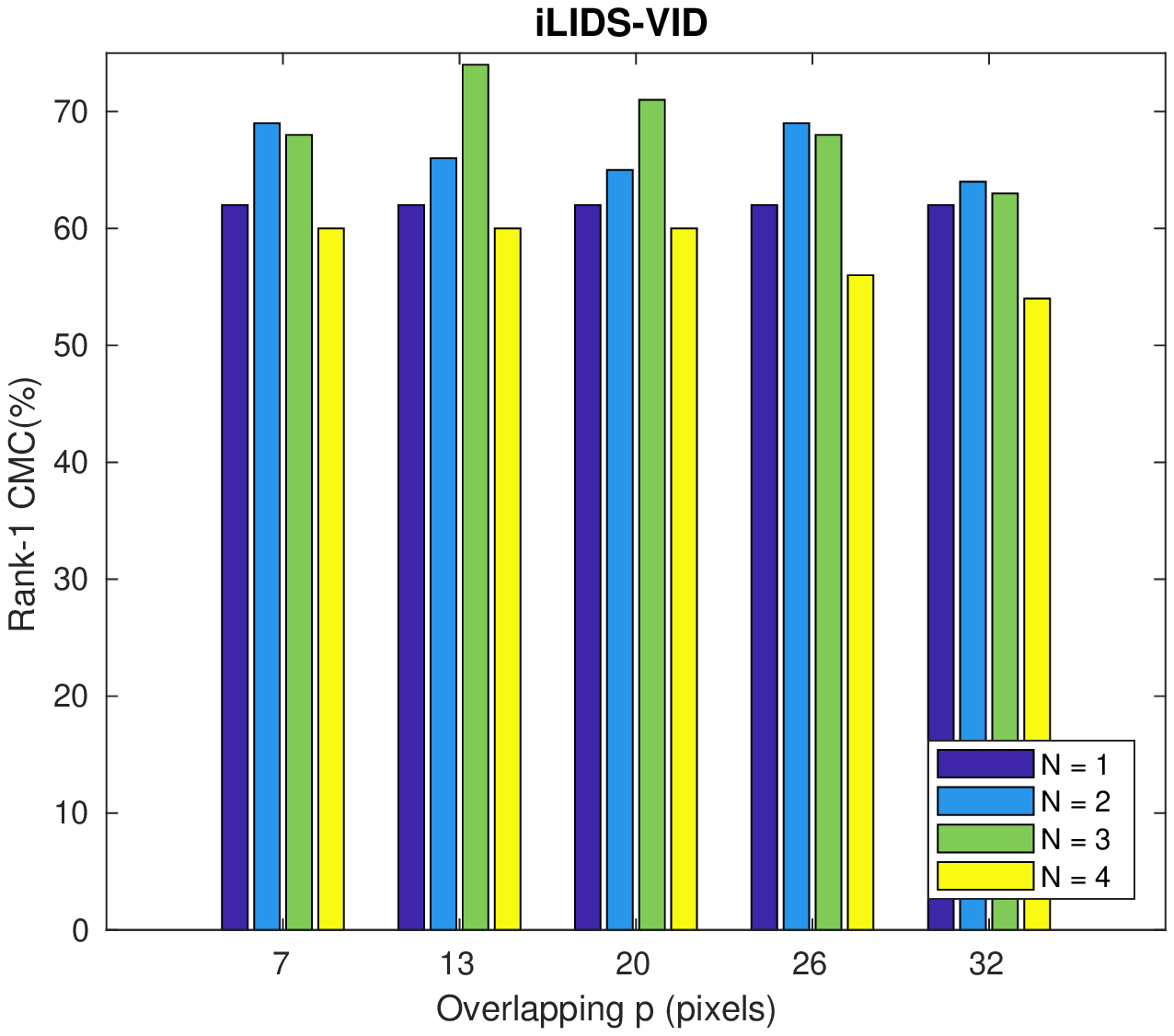}
    \end{minipage}
  }
  \subfigure[]{
    \begin{minipage}{0.48\textwidth}
    \centering
    \includegraphics[width=0.9\textwidth, height=0.7\textwidth]{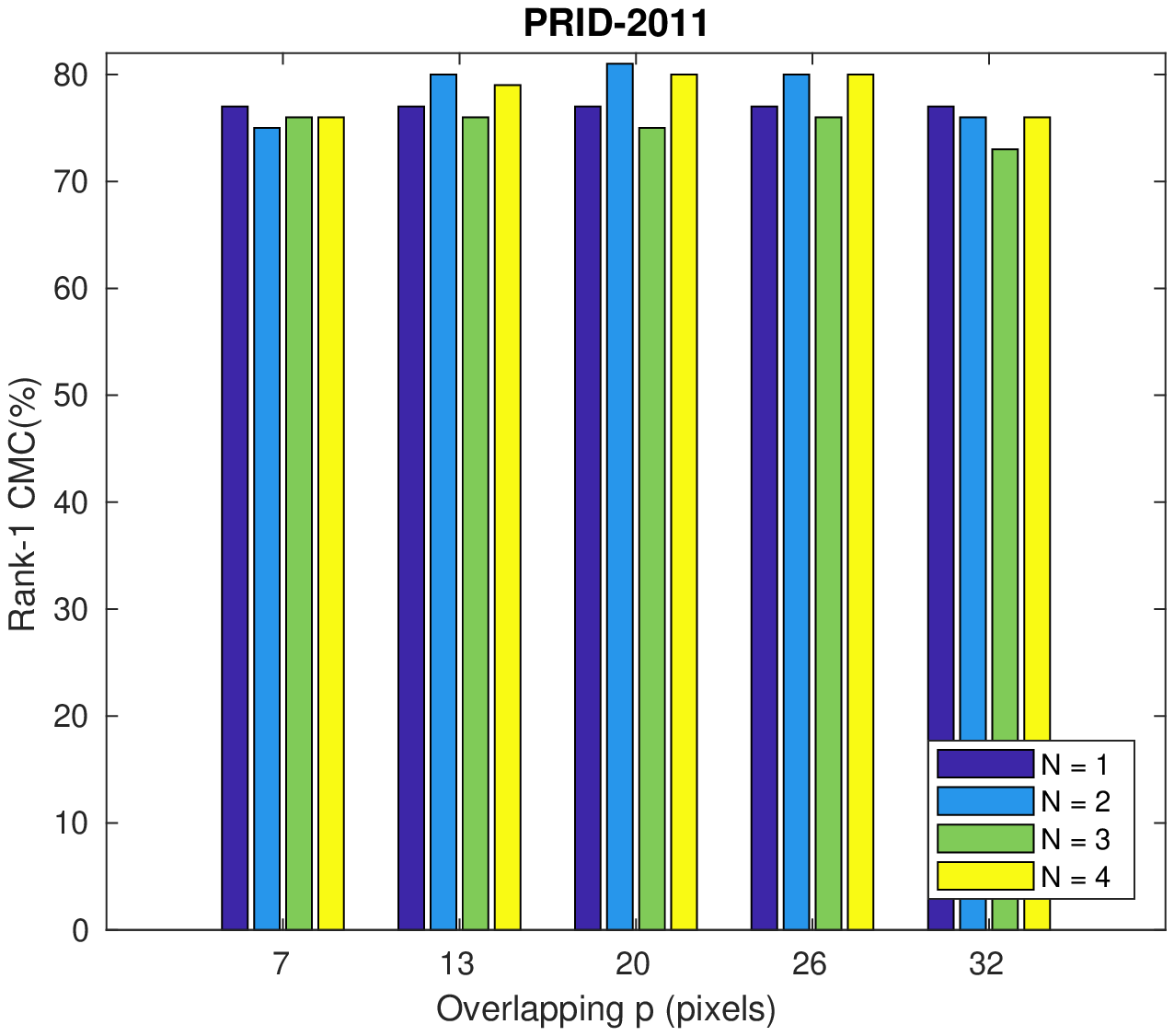}
    \end{minipage}
  }
  \caption{The rank-1 CMC accuracy on the iLIDS-VID and PRID-2011 datasets with the part numbers $N$ and the pixels $p$.}
  \label{figureilidsprid1}
\end{figure}

\subsection{Parts Number, Margin and Overlapping}
\label{npartmarginoverlapping}

In section \ref{netinput}, we split each frame of video sequence in a specific way, in which there are two main parameters: the parts number $N$ and the overlapping pixels $p$. In Eq. \ref{marginequation} of the section \ref{lossfunction}, there is a margin value $m$ which plays an important role in Siamese loss function. In this section, we exploit the effect of the parameters $N$, $p$, and $m$.
We first let the margin $m$ be one of $\{1,2,3,4\}$, through a large number of experiments, we concluded that the STFMM with the margin value $m=2$ performed better in overall CMC accuracies. Therefore we set $m=2$ in the following experiments.

We set the number of parts $N$ as one of $\{1,2,3,4\}$ and split each frame horizontally. The overlapping $p$ of two adjacent parts is also an important parameter, we choose it according to the percentages of the height of the original frame at five levels $5\%,10\%,15\%,20\%$, and $25\%$. In our experiments, the height of the original frame is 128, therefore the value of $p$ is set as one of $\{7,13,20,26,32\}$.

The experimental results of the rank-1 CMC re-identification accuracy with different values of $N$ and $p$ are shown in Fig. \ref{figureilidsprid1}. We can see that the models with the input of the part sequences have better performances compared with the models of using the full frames, and a number of overlapping pixels can improves the accuracy for person re-identification. The model with $N=3$ and $p=13$ outperforms the one with $N=1$ (full frame sequence) by 12\% of rank-1 accuracy on the iLIDS-VID dataset. The model with $N=2$ and $p=20$ outperforms the one with $N=1$ by 4\% of rank-1 accuracy on the PRID-2011 dataset. The results demonstrate that STFMM can significantly improve the accuracy for person re-identification, especially on the complicated dataset iLIDS-VID.
\begin{figure}
  \centering
  \subfigure[]{
    \begin{minipage}{0.48\textwidth}
    \centering
    \includegraphics[width=0.9\textwidth, height=0.7\textwidth]{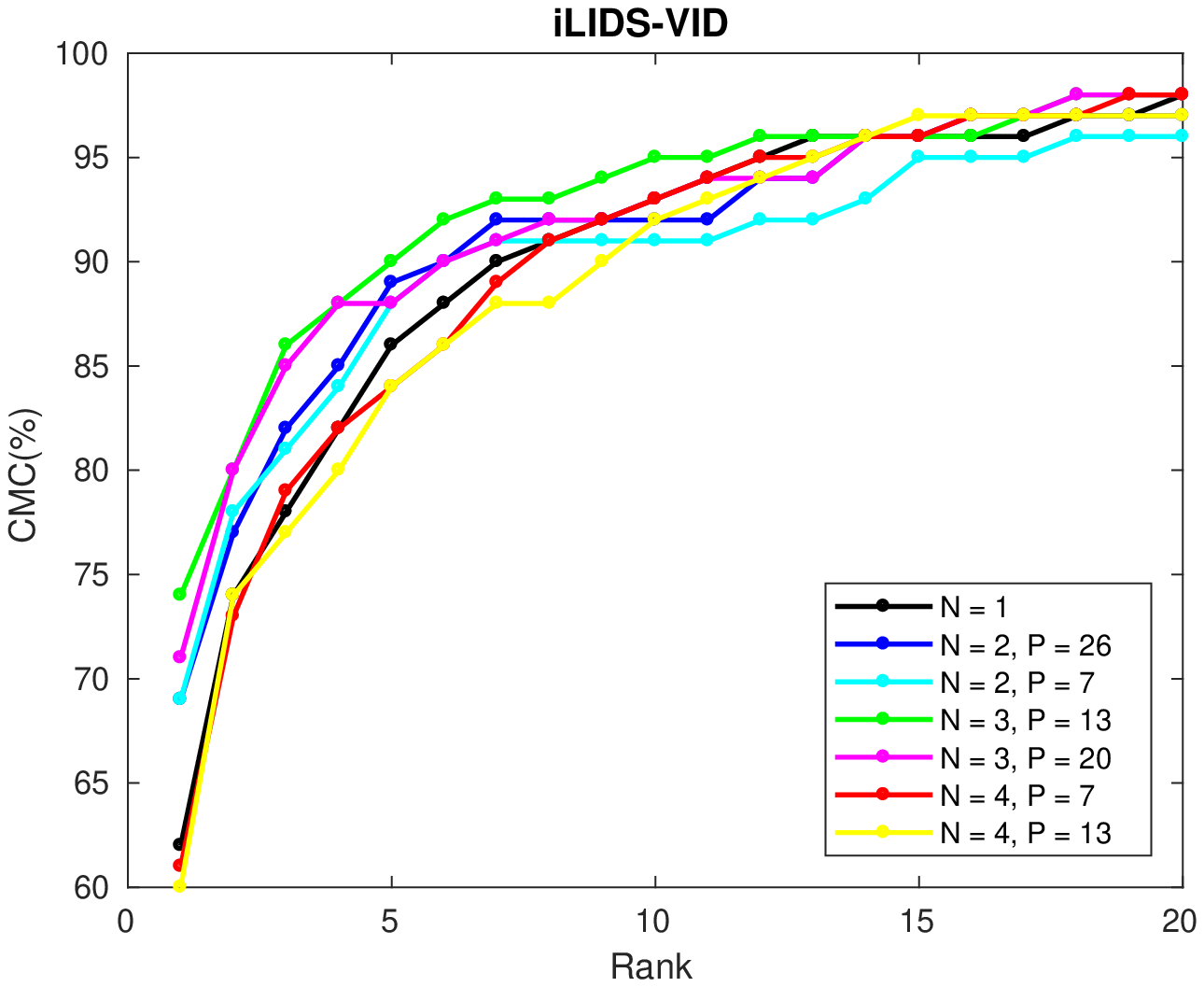}
    \end{minipage}
  }
  \subfigure[]{
    \begin{minipage}{0.48\textwidth}
    \centering
    \includegraphics[width=0.9\textwidth, height=0.7\textwidth]{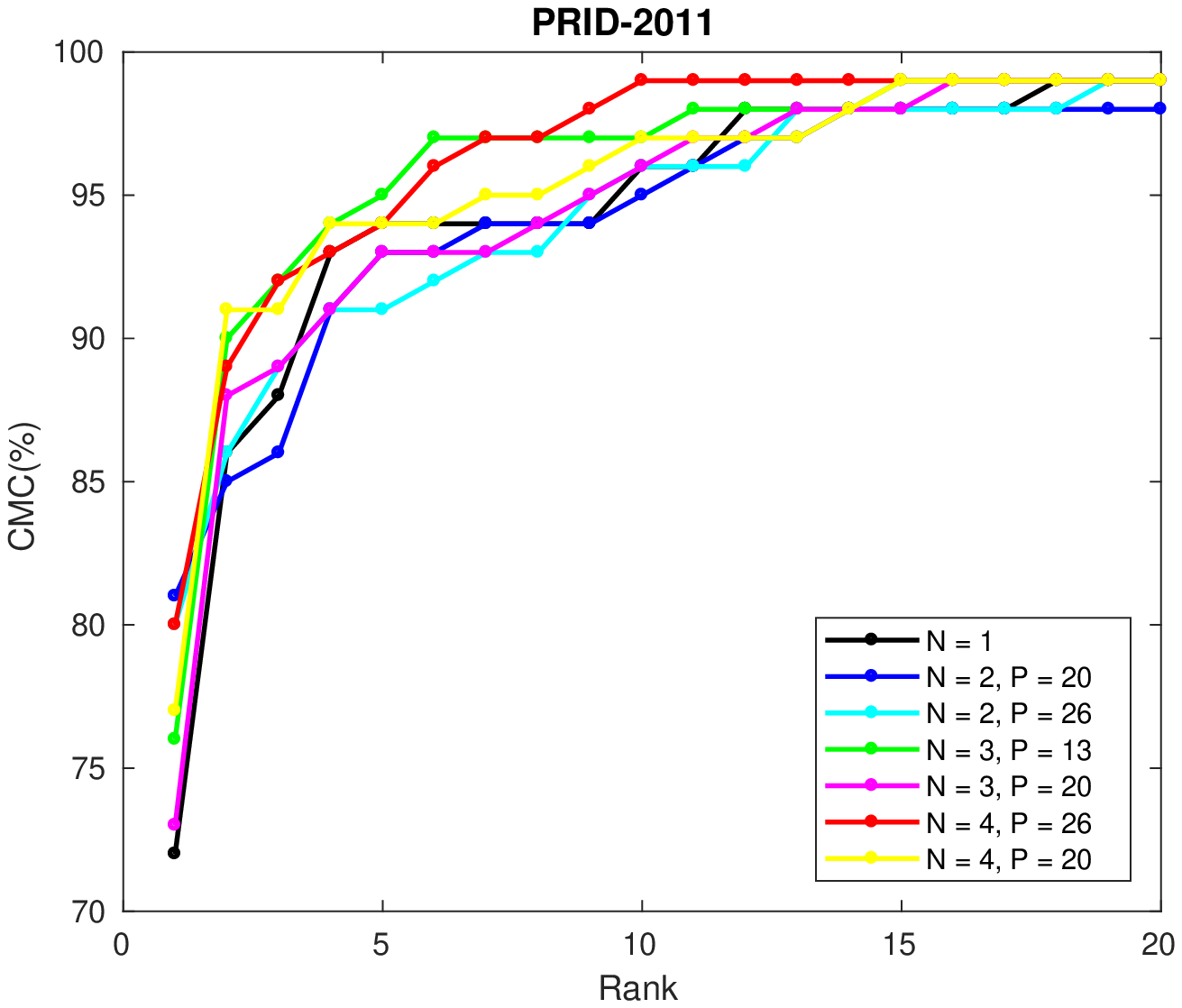}
    \end{minipage}
  }
  \caption{The top two CMC curves of each value of $N\in \{2,3,4\}$ on the iLIDS-VID and PRID-2011 datasets, and the CMC curve of $N=1$ is in the case of full frame sequence.}
  \label{figureilidsprid2}
\end{figure}

 The top two CMC curves of each value of $N\in \{2,3,4\}$ are shown in Fig. \ref{figureilidsprid2}. We can see that STFMM with $N=3$ and $p=13$ works best on the rank-5 CMC accuracy with the accuracy of 90\% on iLIDS-VID, and 94\% on PRID-2011. Therefore if we aim to provide several candidate results for person re-identification, the mode with $N=3$ and $P=13$ would be a good choice. In the experiments of testing the parts numbers $N$, we found that the training time of our model was directly proportional to the value of $N$.
\subsection{Comparison with the State-of-the-art Methods}
Except for the methods DVR \cite{bt10}, CNN+RNN \cite{bt8}, ASTPN \cite{bt12}, and TAM+SRM \cite{bt13} described in Section \ref{relatedwork}, there are several other methods for video-based person re-identification. AFDA \cite{bt11} was an algorithm which can hierarchically cluster video sequences and utilize the representative frames to learn a feature subspace maximizing the Fisher criterion. STA \cite{bt9} utilized the spatio-temporal body-action model to exploit the periodicity exhibited by a walking person and build a spatio-temporal appearance representation for pedestrian re-identification. In RFA \cite{bt3}, Yan \etal proposed a network based on LSTM to aggregate the frame-wise representation of human and yielded a sequence level representation.

The comparison of our STFMM method and the previous methods is presented in Tab. \ref{comparewithstateoftheart}. We can see that our method outperforms the previous methods significantly with the rank-1, rank-5, and rank-10 CMC accuracy of 74\%, 90\% and 95\% respectively on the dataset iLIDS-VID. In particular, it exceed the most recently method ASTPN by 12\% on the rank-1 CMC accuracy. On the dataset PRID-2011, our method improves the rank-1 CMC accuracy from 77\% to 81\% compared with the method ASTPN. The results demonstrate that our method performs better than the methods of using the input of full body sequences.
\begin{table}
  \footnotesize
  \centering
  \begin{tabular}{c|c|c|c|c|c|c|c|c}
    \hline
    Datasets & \multicolumn{4}{|c|}{iLIDS-VID} & \multicolumn{4}{|c}{PRID-2011} \\
    \hline 
    CMC Rank            & R=1 & R=5 & R=10 & R=20 & R=1 & R=5 & R=10 & R=20 \\
    \hline 
    AFDA \cite{bt11}    & 38  & 63  & 73   & 82   & 43  & 73  & 85   & 92  \\
    DVR \cite{bt10}      & 35  & 57  & 68   & 78   & 42  & 65  & 78   & 89  \\
    STA \cite{bt9}      & 44  & 72  & 84   & 92   & 64  & 87  & 90   & 92  \\
    RFA \cite{bt3}      & 49  & 77  & 85   & 92   & 64  & 86  & 93   & 98  \\
    RNN+CNN \cite{bt8}  & 58  & 84  & 91   & 96   & 70  & 90  & 95   & 97  \\
    TAM+SRM \cite{bt13} & 56  & 86  & 92   & 97   & 80  & 95  & 99   & 99  \\
    ASPTN \cite{bt12}   & 62  & 86  & 94   & \textbf{98}   & 77  & \textbf{95}  & \textbf{99}   & \textbf{99}  \\
    STFMM              & \textbf{74}  & \textbf{90}  & \textbf{95}   & \textbf{98}   & \textbf{81}  & 94  & \textbf{99}   & \textbf{99}  \\
    \hline
  \end{tabular}
  \caption{ Comparison of STFMM method with previous state-of-the-art methods on iLIDS-VID and PRID-2011 in terms of Rank CMC(\%).}
  \label{comparewithstateoftheart}
\end{table}
\begin{table}
  \footnotesize
  \centering
  \begin{tabular}{c|c|c|c|c|c}
    \hline
    Model & Trained on & R=1 & R=5 & R=10 & R=20 \\
    \hline
    RNN+CNN \cite{bt8} & iLIDS-VID & 28 & 57 & 69 & 81  \\
    ASPTN \cite{bt12}  & iLIDS-VID & 30 & 58 & 71 & 85  \\
    STFMM             & iLIDS-VID & \textbf{48} & \textbf{77} & \textbf{90} & \textbf{94}  \\
    \hline
  \end{tabular}
  \caption{ Cross-dataset testing accuracy tested on PRID-2011 in terms of Rank CMC (\%).}
  \label{crossdatatesting}
\end{table}
\subsection{Cross-Dataset Testing}
Considering that one model may be over-fitting to a particular scenario, cross-dataset testing would be a better way to evaluate its stability, in which the model is trained on the dataset A and tested on the different dataset B. We perform the cross-dataset testing by using 50\% persons of iLIDS-VID for training and 50\% persons of PRID-2011 for testing. Referring to the conclusion of section \ref{npartmarginoverlapping}, our model is trained with $N=3$, $p=13$ and $m=2$. As presented in Tab. \ref{crossdatatesting}, STFMM outperforms the previous methods significantly with the rank-1, rank-5, rank-10 and rank-20 CMC accuracy of 48\%, 77\%, 90\% and 94\% respectively, which exceeds the most recently method ASTPN by 18\%, 19\%, 19\% and 9\% on the corresponding levels. It can be concluded that our model is more stable in practical applications.
\section{Conclusion}
In this paper, we propose a novel deep neural network architecture with spatial and temporal features mixture model (STFMM) for video-based person re-identification. Different from the previous methods, we first split the human body to $N$ parts in horizontal direction in order to obtain more specific information, and the adjacent parts are overlapped with $p$ pixels. In order to satisfy the input of our model, we develop the Siamese network architecture in which the number of input are adjusted from $2$ to $2N$. After choosing the appropriate values of the parameters $m$, $N$ and $p$, we evaluate our model on the iLIDS-VID and PRID-2011 datasets. The experimental results demonstrate that our approach outperforms the existing methods for video-based person re-identification. Specifically, it achieves a rank-1 CMC accuracy of 74\% on the iLIDS-VID dataset, exceeding the most recently developed method ASTPN by 12\%. In order to evaluate the stability of our model, we do the cross-data testing which is trained on the iLIDS-VID dataset and tested on the PRID-2011 dataset. The results show that our model achieves the rank-1, rank-5, and rank-10 CMC accuracies of 48\%, 77\%, 90\% respectively which exceeds ASTPN by no less than 18\% at three corresponding levels. In future, we consider to apply our method to real target tracking or detection system.

\subsubsection*{Acknowledgments}

This research was supported by the National Natural Science Foundation of China (NSFC 61572005, 61672086, 61702030, 61771058), and Key Projects of Science and Technology Research of Hebei Province Higher Education [ZD2017304].

\bibliography{egbib}
\end{document}